\documentclass[sigconf, screen]{acmart}

\AtBeginDocument{%
 }

\copyrightyear{2023}
\acmYear{2023}
\setcopyright{rightsretained}
\acmConference[KDD '23]{Proceedings of the 29th ACM SIGKDD Conference on Knowledge Discovery and Data Mining}{August 6--10, 2023}{Long Beach, CA, USA}
\acmBooktitle{Proceedings of the 29th ACM SIGKDD Conference on Knowledge Discovery and Data Mining (KDD '23), August 6--10, 2023, Long Beach, CA, USA}
\acmDOI{10.1145/3580305.3599846}
\acmISBN{979-8-4007-0103-0/23/08}

\makeatletter
\gdef\@copyrightpermission{
  \begin{minipage}{0.3\columnwidth}
   \href{https://creativecommons.org/licenses/by/4.0/}{\includegraphics[width=0.90\textwidth]{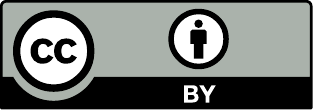}}
  \end{minipage}\hfill
  \begin{minipage}{0.7\columnwidth}
   \href{https://creativecommons.org/licenses/by/4.0/}{This work is licensed under a Creative Commons Attribution International 4.0 License.}
  \end{minipage}
  \vspace{5pt}
}
\makeatother

\usepackage{amsmath,amsfonts,bm}

\def\Figref#1{Figure~\ref{#1}}

\def\Secref#1{Section~\ref{#1}}

\def\eqref#1{equation~\ref{#1}}

\def\Algref#1{Algorithm~\ref{#1}}

\def\1{\bm{1}}

\def\vg{{\bm{g}}}

\def\vr{{\bm{r}}}

\def\vw{{\bm{w}}}

\def\mG{{\bm{G}}}

\DeclareMathAlphabet{\mathsfit}{\encodingdefault}{\sfdefault}{m}{sl}
\SetMathAlphabet{\mathsfit}{bold}{\encodingdefault}{\sfdefault}{bx}{n}

\def\sR{{\mathbb{R}}}

\newcommand{\norm}[1]{\left\lVert#1\right\rVert}

\usepackage[utf8]{inputenc} %
\usepackage[T1]{fontenc}    %
\usepackage{amsfonts}       %

\usepackage{multirow}
\usepackage{bm}
\usepackage{algorithm,algorithmic}
\usepackage{subcaption}
\usepackage{wrapfig}
\usepackage[capitalise]{cleveref}

\begin{document}

\title{Improving Training Stability for Multitask Ranking Models in Recommender Systems}

\author{Jiaxi Tang$^*$}
\affiliation{%
  \institution{Google Deepmind}
  \city{Mountain View}
  \state{California}
  \country{USA}}
\email{jiaxit@google.com}
 
\author{Yoel Drori$^*$}
\affiliation{%
  \institution{Google Research}
  \city{Tel Aviv}
  \country{Israel}}
\email{dyoel@google.com}

\author{Daryl Chang$^*$}
\affiliation{%
  \institution{Google Inc}
  \city{Mountain View}
  \state{California}
  \country{USA}}
\email{dlchang@google.com}

\author{Maheswaran Sathiamoorthy}
\affiliation{%
  \institution{Google Deepmind}
  \city{Mountain View}
  \state{California}
  \country{USA}}
\email{nlogn@google.com}

\author{Justin Gilmer}
\affiliation{%
  \institution{Google Deepmind}
  \city{Mountain View}
  \state{California}
  \country{USA}}
\email{gilmer@google.com}

\author{Li Wei}
\affiliation{%
  \institution{Google Inc}
  \city{Mountain View}
  \state{California}
  \country{USA}}
\email{liwei@google.com}

\author{Xinyang Yi}
\affiliation{%
  \institution{Google Deepmind}
  \city{Mountain View}
  \state{California}
  \country{USA}}
\email{xinyang@google.com}

\author{Lichan Hong}
\affiliation{%
  \institution{Google Deepmind}
  \city{Mountain View}
  \state{California}
  \country{USA}}
\email{lichan@google.com}

\author{Ed H. Chi}
\affiliation{%
  \institution{Google Deepmind}
  \city{Mountain View}
  \state{California}
  \country{USA}}
\email{edchi@google.com}
\renewcommand{\shortauthors}{Jiaxi Tang et al.}

\begin{abstract}
Recommender systems play an important role in many content platforms. While most recommendation research is dedicated to designing better models to improve user experience, we found that research on stabilizing the training for such models is severely under-explored. As recommendation models become larger and more sophisticated, they are more susceptible to training instability issues, \emph{i.e.}, loss divergence, which can make the model unusable, waste significant resources and block model developments. In this paper, we share our findings and best practices we learned for improving the training stability of a real-world multitask ranking model for YouTube recommendations. We show some properties of the model that lead to unstable training and conjecture on the causes. Furthermore, based on our observations of training dynamics near the point of training instability, we hypothesize why existing solutions would fail, and propose a new algorithm to mitigate the limitations of existing solutions. Our experiments on YouTube production dataset show the proposed algorithm can significantly improve training stability while not compromising convergence, comparing with several commonly used baseline methods. We open source our implementation at {\footnotesize \url{https://github.com/tensorflow/recommenders/tree/main/tensorflow_recommenders/experimental/optimizers/clippy_adagrad.py}}.
\end{abstract}

\ccsdesc[500]{Information systems~Recommender systems}
\ccsdesc[500]{Information systems~Retrieval models and ranking}

\keywords{Recommender System; Optimization; Training Stability}

\maketitle

\def\thefootnote{*}\footnotetext{Equal contribution to the work.}\def\thefootnote{\arabic{footnote}}

\section{Introduction}\label{sec:intro}
A good recommender system plays a key factor to user experience. It has become a core technology and even a main user interface in many web applications, including YouTube, one of the largest online video platforms in the world.
As a result, many components can be incorporated into recommendation models to capture contexts with different modalities and improve recommendation quality, including audio signals~\cite{van2013deep}, video signals~\cite{lee2020large}, user history sequence~\cite{beutel2018latent,tang2019towards}, etc. Besides, the scaling law of recommendation models~\cite{ardalani2022understanding} suggests substantial quality improvements by increasing model capacity in data-rich applications.

As recommendation models become larger and more sophisticated, they are more susceptible to training instability issues~\cite{gilmer2021loss}, \emph{i.e.}, the loss diverges (instead of converging), causing the model to be ``broken’’ and completely useless.
In industry, serving such a ``broken’’ model leads to catastrophic user experience (see~\Secref{sec:background_motivation}). Moreover, if we cannot ensure reliable training of recommendation models, a huge amount of resources can be wasted and model development can be blocked. Therefore, we couldn't emphasize more on how essential training stability is.
However, very sparse research has been done on the training stability of recommendation models. 

On one hand, there's a lack of fundamental understanding of why recommendation models are prone to training instability issues. In particular, we observe that ranking models with multiple objectives are more likely to encounter problems than retrieval models with a single objective (\emph{e.g.} Softmax Cross-Entropy over large output space). In addition to increasing model complexity, we found that simply adding new input features or output tasks can also cause training unstable.
To deal with this problem, people mostly rely on empirical solutions and sometimes on luck (when the problem occurs randomly).
Developing a fundamental understanding of what causes the problem would allow people to navigate the process more confidently.

On the other hand, we found that there's a lack of effective approaches to largely mitigate the training instability problem.
There are some widely used methods, such as activation clipping~\cite{krizhevsky2010convolutional}, gradient clipping~\cite{pascanu2013difficulty, brock2021high}, learning rate warmup~\cite{gilmer2021loss,cohen2021gradient}, and layer normalization~\cite{ba2016layer}. But in practice, we found that these approaches were ad hoc and couldn't completely prevent training instability in our model.
Developing an effective method that can significantly improve model training stability accelerates model improvements by addressing concerns about training problems.

The focus of this paper is to share the lessons learned from addressing the training instability problems experienced by a multi-task ranking model used in production for YouTube recommendations. 
In~\Secref{sec:background}, we show some implications and consequences of unstable model training in real-world recommender systems, suggesting the importance and difficulties of considerably improving model training stability.
After introducing some preliminary basics of our model in~\Secref{sec:understand}, we present some case studies about changes that had led to more training instability problems and provide our understanding on the root cause of the problems.
In practice, however, we've found that there's a big gap between knowing the root cause and having an effective solution. Some methods that are supposed to be effective do not work well empirically.
Next, in~\Secref{sec:method}, we closely examine the training dynamics of our model, which inspired us to propose a more effective approach to overcome the limitations in existing methods.
The empirical evidence on a YouTube dataset in~\Secref{sec:experiments} reveals the effectiveness of the proposed method for improving model training stability, especially when increasing model capacity and using a large learning rate for faster convergence. 
We hope that these findings can help the community better understand the training instability problem and effectively solve it.

\section{Background and Related Work}\label{sec:background}
\begin{figure}[t!]
\centering
\includegraphics[scale=0.37]{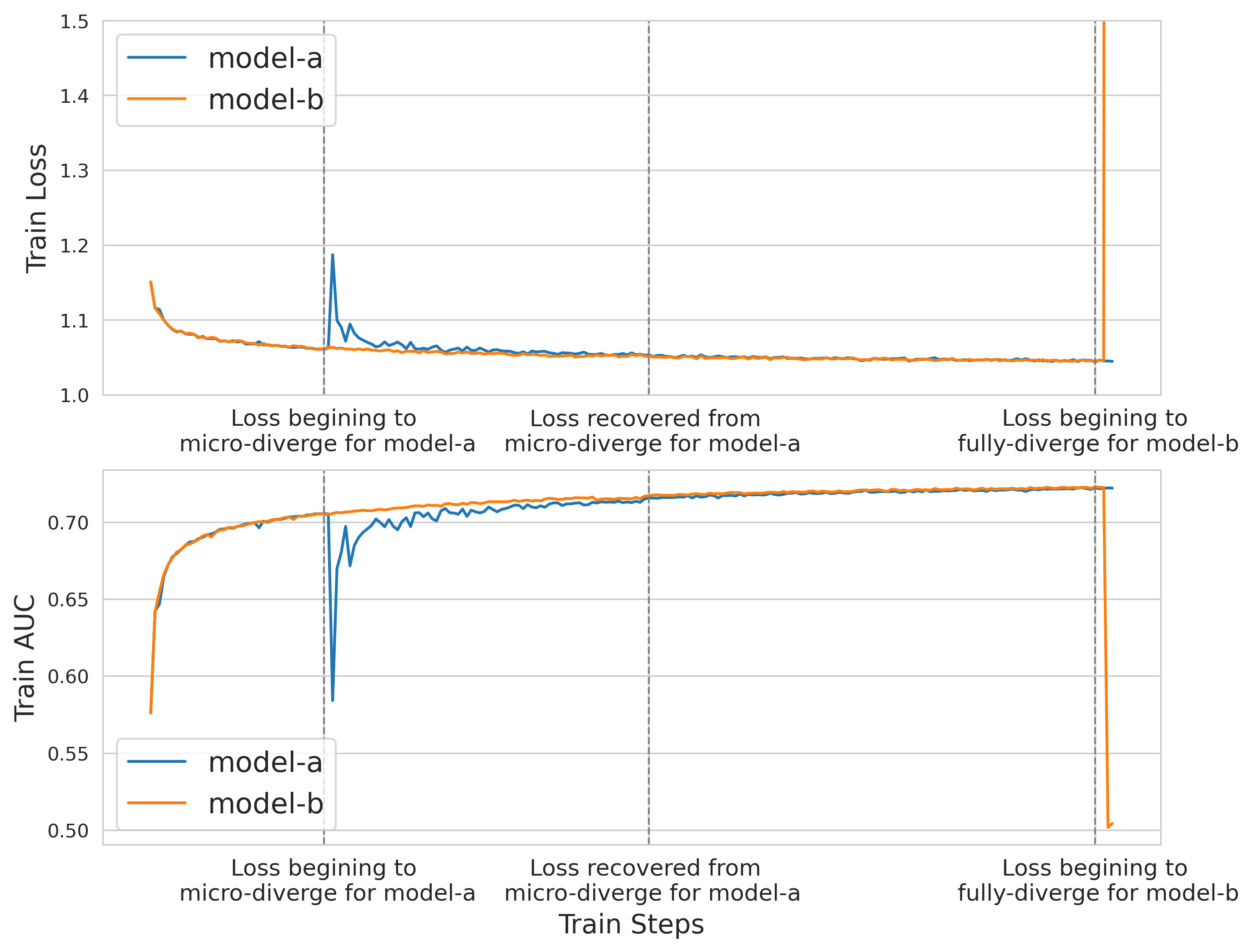}
\caption{Example of loss divergence in our model and its impact on training loss (top) and AUC (bottom). In this example, model-a's loss micro-diverged then recovered, whereas model-b's loss fully-diverged.}
\label{fig:divergence}
\end{figure}

\subsection{Symptoms}
Training instability is a model property that measures the unsteadiness of model training. It has a common symptom of \emph{loss divergence} (\emph{a.k.a} loss blow-up).
Based on our observations, we further categorize loss divergence into two types: \emph{micro-divergence} and \emph{full divergence}. 
When a model's loss micro-diverges (see \emph{model-a} in~\Figref{fig:divergence} as an example), we can observe a sudden jump in training loss and a sudden drop in training metrics, although the loss may recover to normal (as shown in the example) as training continues. Usually, we don't need to worry too much about this situation, because the recovered model can have a quality on-par with models that don't suffer from loss divergence.
However, if a model's loss fully diverges (see \emph{model-b} in~\Figref{fig:divergence} as an example), we can see that the training loss becomes very high in a few number of training steps, and all training metrics become extremely bad. For example, the binary classification AUC (the metric we mainly look throughout the paper) drops to 0.5 as shown in~\Figref{fig:divergence}, suggesting the model becomes completely useless, practically giving random results. What’s worse, the fully diverged loss cannot recover to its pre-divergence value as training continues.

\subsection{Motivation and Challenges}\label{sec:background_motivation}
We motivate the importance of training stability research, especially for recommender systems in industry, from several aspects.
First, the problem of loss divergence, once it occurs regularly, can affect almost all types of model development. This includes, but is not limited to: 
\begin{enumerate}
\item \textbf{Increasing model complexity:} 
As more modeling techniques are applied and more components are added to the recommendation model (to improve its quality), there's a greater chance that the model will suffer from loss divergence problems. Even simply enlarging the model capacity could put the model in a dangerous state, despite the great benefits in data-rich environments suggested by current scaling laws~\cite{ardalani2022understanding}.

\item \textbf{Adding more input features or tasks:} 
Typically, the ranking model in a recommendation system uses many input features for multiple tasks~\cite{zhao2019recommending}. A combination of predictions on all tasks is used to decide the ranking of a candidate item. We found that both adding new input features and adding new tasks can lead to training instability, although they are common ways to improve model quality.

\item \textbf{Increasing convergence speed:} 
We have found that hyperparameter tuning that facilitate model convergence (such as increasing the learning rate) can significantly increase the likelihood of loss divergence. This forces model designers to use a smaller learning rate which results in slower convergence.
\end{enumerate}
Second, as training complex models requires large amounts of resources, loss divergence problems, which block the model from completing their training, waste training resources. Moreover, inadvertently deploying a ``broken'' model for serving also leads to catastrophic user experience.

Consequently, we've seen many efforts on alleviating this problem from an engineering perspective, such as ensuring model quality before serving. Nevertheless, given that engineering efforts cannot prevent training instability from occurring, it is clear that drastically improving model training stability is the right path to pursue in the long run.

In dealing with the problem of model instability, we experienced the following challenges.
\begin{itemize}
\item \textbf{Reproducibility}: 
A model can suffer from loss divergence at any time during training, however, only some of them can be easily reproduced (more discussions in~\Secref{sec:understand}). Failing to reproduce bad cases makes it hard to understand what happened to the model before loss divergence.

\item \textbf{Detection:} 
In practice, it is costly to evaluate model and report results frequently during training, otherwise the training can be significantly slowed down. Since sometimes a micro-divergence can happen and then recover very quickly, it is hard to even detect if any micro-divergence happens during model training without sacrificing training speed.

\item \textbf{Measurement:} 
There's little research on quantitative measures of a model's training stability prior to training. To know (1) whether a modeling change will increase the risk of loss divergence, or (2) whether a mitigation can help reduce the risk of loss divergence, one has to rely on empirical evaluations (\emph{i.e.,} train multiple copies of a model and check how many of them have issues), which are time-consuming and resource-intensive.
\end{itemize}

\subsection{Related Work}
Model training stability has been an under-explored research area, not only for recommendation models, but also in general machine learning. Fortunately, with the increasing trend of large models~\cite{thoppilan2022lamda, chowdhery2022palm, brown2020language}, stabilizing model training has become an emerging research area and attracts more attention in recent years.

From the perspective of optimization theory, \citet{wu2018sgd} first theoretically predicted the training instability for quadratic models with learning rate and the ``sharpness'' (measured by the maximum eigenvalue of the loss Hessian) of the loss curvature. For deep neural networks, \citet{cohen2021gradient, gilmer2021loss} confirmed that this prediction is still accurate enough.

In terms of techniques, there are some methods widely used in language and vision models, such as activation clipping~\cite{krizhevsky2010convolutional}, gradient clipping~\cite{pascanu2013difficulty}, learning rate warmup~\cite{goyal2017accurate}, and various normalization techniques~\cite{ioffe2015batch, ba2016layer}. In addition, \citet{you2019large} proposed a new optimizer that achieves a better trade-off between convergence and stability for large batch-size training. \citet{brock2021high} developed Adaptive Gradient Clipping to improve the stability of ResNet models~\cite{he2016deep} without Batch Normalization~\cite{ioffe2015batch}. 

However, empirically, we found these approaches not effective enough to completely prevent our model from training instability (See~\Secref{sec:experiments}). This may due to some unique properties of the recommendation model. As will be discussed in the next section, these properties can make multi-task ranking models more susceptible to training instability problems.

\section{Understanding the Cause of the Issue}\label{sec:understand}

\begin{figure}[t!]
\centering
\includegraphics[scale=0.5]{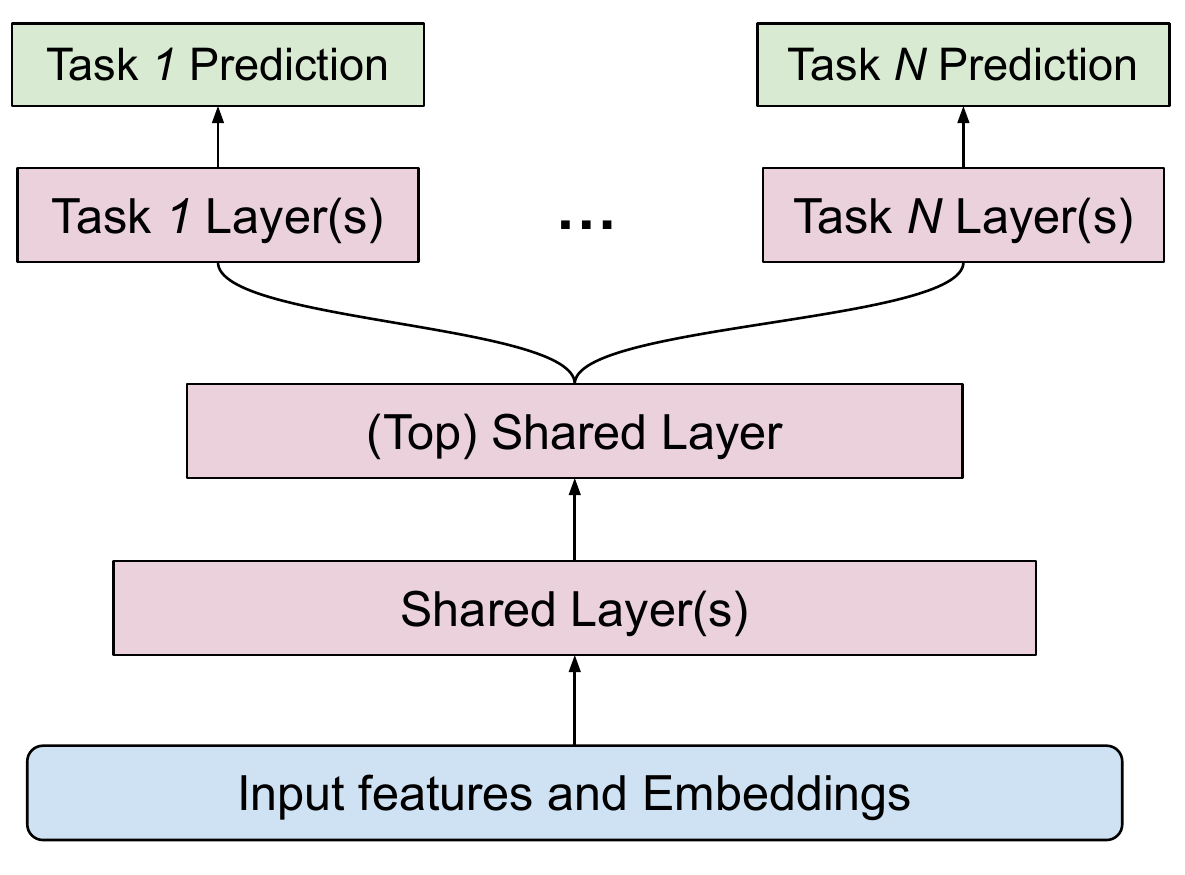}
\caption{An general illustration of the ranking model used in recommender systems. The model has one or more layers that are (softly or fully) shared by multiple tasks.}
\label{fig:model}
\end{figure}

In this section, we first describe the model to be studied in this paper and its characteristics. Then, we share our understanding on the root cause of the training instability problems that happened in our model.

\subsection{Model Definition}\label{sec:understand_model}
YouTube's video recommendation system uses multiple candidate generation algorithms to retrieve a few hundred candidates. This is followed by a ranking system which generates a ranked list from these candidates.
This paper mainly focuses on the ranking models in YouTube's recommender system. Different from candidate generation models (\emph{a.k.a} retrieval models), which are responsible for filtering out the majority of irrelevant items, ranking models aim to provide a ranked list so that items with the highest utility to users are displayed at the top. 
Therefore, ranking models use more advanced machine learning techniques with more expensive features to have sufficient model expressiveness for learning the association of features and their relationship with utility.

\Figref{fig:model} depicts a general architecture of the ranking model that we want to study throughout the paper. Below we summarize some important features for our ranking model and how it is trained; one can refer to~\cite{zhao2019recommending} for more details.

\begin{itemize}
    \item \textbf{Multitask:}
    As shown in \Figref{fig:model}, the ranking model has multiple tasks that predict multiple labels. These predictions are combined to form the final ranked list of items. Regardless of different modeling choices~\cite{ma2018mtr, caruana1997multitask}, there are some hidden layers in the middle of the model that are shared by these tasks (either fully shared or softly shared).
    
    \item \textbf{Sequential training:}
    The model is trained sequentially, \emph{i.e.,} the training is done over a corpus of data in sequential order from old to new. While unlike pure online learning~\cite{bottou2003large,shalev2012online}, which visits training data in strictly sequential order, we define a time-based moving window and randomly sample data batches from this window of training data for parallel training. This training scheme has been widely used and is known to be beneficial for many aspects of recommendation quality~\cite{zhao2019recommending,mcmahan2013ad, anil2022factory}.
    
    \item \textbf{Optimization:}
    Large batch-size training is known to have less noise in gradients and thus optimization is more curvature driven~\cite{you2019large,anil2022factory}. We adopt a large batch size with a high learning rate for faster convergence. We found Adagrad~\cite{duchi2011adaptive} to be strong in our case, despite many advances in optimizers (e.g., Adam~\cite{kingma2014adam}, Adafactor~\cite{shazeer2018adafactor}).
\end{itemize}

\subsection{Root Cause and Case Studies}\label{sec:understand_cause}

\begin{figure}[t!]
\centering
\includegraphics[scale=0.4]{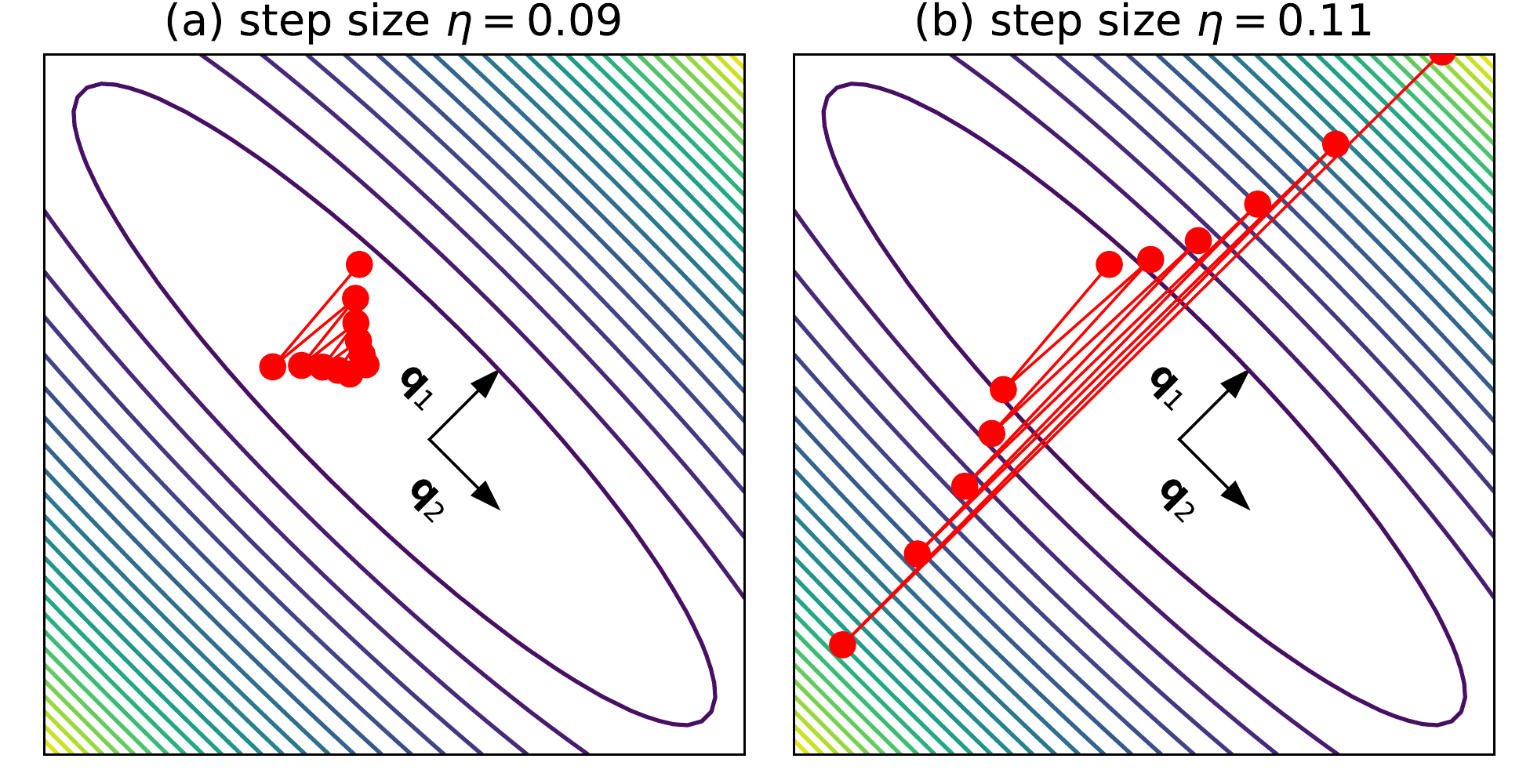}
\caption{From~\cite[Figure~2]{cohen2021gradient}. Gradient descent on a quadratic model with eigenvalues $\alpha_1=20$ and $\alpha_2=1$. We can clearly observe training instability problems starting to occur when learning rate $\eta > 2/\alpha_* = 2/\alpha_1 = 0.1$.}
\label{fig:quadratic}
\end{figure}

Regardless of the types of loss divergence, we believe that the intrinsic cause can be summarized as \textbf{``step size being too large when loss curvature is steep''}.
Once a model meets both conditions at a given state, a divergence can easily occur. Intuitively, the step size should be conservative at a steep loss surface (measured by the maximum eigenvalue of the loss Hessian) to ensure that loss decreases instead of increases. 

For quadratic models, \citet{wu2018sgd} theoretically proves the above argument and suggests
$$
2 / \eta > \alpha_*
$$
to make training stable, where $\eta$ is the learning rate and $\alpha_*$ is the maximum eigenvalue of the loss Hessian. \citet{cohen2021gradient} gives a nice and straightforward example (in \Figref{fig:quadratic}) for the proof. For neural networks, this argument still mostly holds~\cite{cohen2021gradient, gilmer2021loss}.

Knowing the root cause of the training instability problem allows us to answer the following research questions:\\

\noindent{\textit{RQ1: Why do recommendation models in general have worse training stability than models in other domains?}}\\

\noindent{\textit{RQ2: Within recommendation models, why do ranking models typically have worse training stability than retrieval models?}}\\

\noindent{We relate the answers to these questions to the following unique properties of our models. Please refer to some empirical evidence in~\Secref{sec:suppl_more_evidence} in Supplementary Material.}
\begin{itemize}
    \item \textbf{Data distribution changes (RQ1)}: 
    Compared to models in other domains, recommendation models use several orders of magnitude more input features (hundreds to thousands). What's worse, with sequential training, the distribution of these input features (and labels) keeps changing. We think a steeper loss curvature can occur when data distribution sudden change, which happens regularly. Also, a model with sequential training will never converge as it has to adapt to newly arriving data points with a changed distribution. Thus, a large learning rate is required to make the adaptation efficient enough. In summary, compared to models in other domains that are trained on a fixed dataset, changes in the training data distribution pose greater challenges for stabilizing the training of recommendation models.
    
    \item \textbf{Larger model size and complexity (RQ2)}: 
    Compared to retrieval models used for candidate generation, ranking models are usually much larger in capacity to accurately measure the utility of candidate items. With the recent developments of ML hardware (\emph{e.g.,} TPUs), we are able to significantly increase the model size for quality improvements~\cite{ardalani2022understanding}.  The empirical studies from~\citet{gilmer2021loss} suggested the increased model capacity and complexity is a contributing factor to steeper loss curvature. 
    
    \item \textbf{Multiple objectives vs. Single objective (RQ2)}: 
    Compared to retrieval models which usually has a single objective (e.g. Softmax Cross-Entropy)~\cite{yi2019sampling}, ranking models often need to optimize for many objectives at the same time~\cite{zhao2019recommending}. This causes ranking models to suffer from loss divergence much more easily. Because if there are spurious gradients caused by bad predictions from a particular task, the gradients can back-propagate throughout the model, causing the layers that are shared by multiple tasks to behave (slightly) abnormally. But since the layers are shared by different tasks, other tasks tend to predict irregular values afterwards, reinforcing the instability to a nonrecoverable state. In other words, shared layers (as well as embeddings) can be a double-edged sword –- they allow transfer learning from different tasks, but can also exacerbate the training instability problem, making ranking models more vulnerable than retrieval models.
    
\end{itemize}

Despite the recent advances in understanding the root causes of divergence issues, we have found a large gap remains between our current understanding on the cause of the issue and having an effective solution. We have tried many temporary fixes. Some examples are: (1) Using even slower learning rate warmup schedule to pass the initial model state where loss curvature is steep~\cite{gilmer2021loss}. (2) Enlarging the sequential training moving window to make training data distribution changes smoother. These fixes indeed mitigated training instability issues for a while, but when our model became more complex, loss divergence happened again.
After trying many ad-hoc fixes, we believe developing a more principled way that can significantly improve model stability is the long-term solution.

\section{Effective method for improving training stability}\label{sec:method}
\begin{figure*}[t!]  %
\centering
    \centering
    \begin{subfigure}[b]{0.95\textwidth}
        \includegraphics[width=\textwidth]{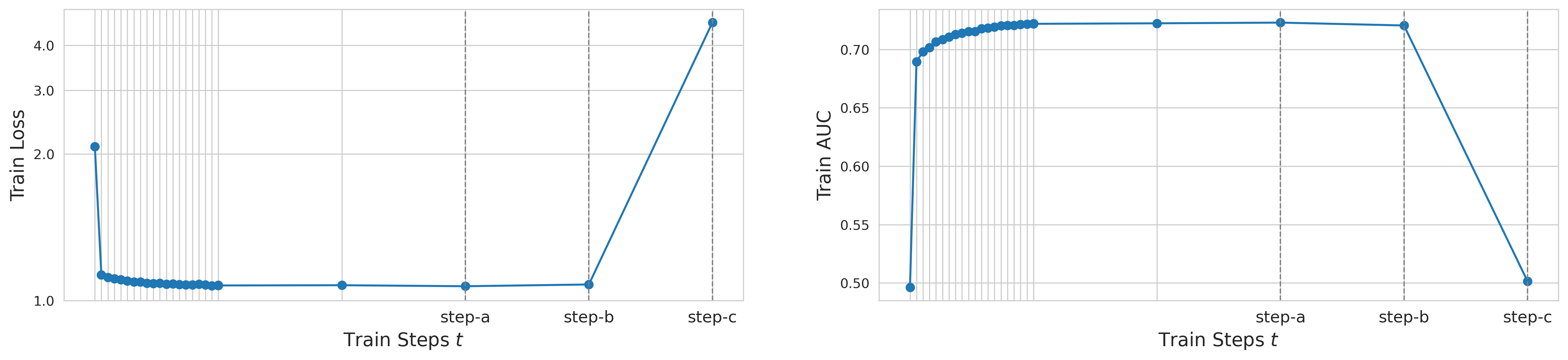}
        \caption{Training loss (left) and AUC (right) at different steps.}
        \label{fig:train_loss_auc}
    \end{subfigure}
    
    \begin{subfigure}[b]{0.95\textwidth}
        \includegraphics[width=\textwidth]{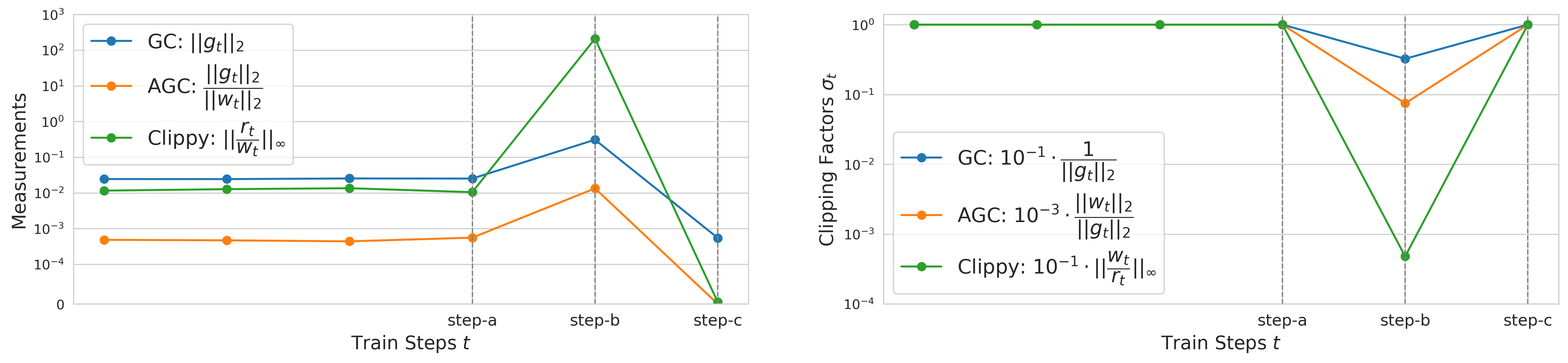}
        \caption{(left) Measurements used by different methods to determine clipping factors. (right) The corresponding clipping factors.}
        \label{fig:norms_factors_top_hidden}
    \end{subfigure}

\caption{(a) We dive into three typical moments in model training: The model was training healthily before step-a. Then at step-b, model's loss aroused and AUC dropped. Finally at step-c, the loss is fully diverged and AUC dropped to 0.5. (b) When checking some statistics from the top hidden layer of the model, we found that GC and AGC failed to provide small enough clipping factor. While Clippy's clipping factor can be 2-orders of magnitude smaller than GC and AGC. \Secref{sec:suppl_statistics} in Supplementary Material has the statistics for other layers.}
\label{fig:training_statistics}
\end{figure*}

In this section, we first introduce the general direction (Gradient Clipping) for controlling the effective step size while loss curvature is steep, by presenting some classical methods on this direction, accompanied with notations and denotations. Despite being successful when applied in other domains, we found these classical methods are not effective enough when applied in our model. Based on some observations of training dynamics in our model, we propose a new method and explain why it can be more effective for improving the training stability.

We first describe Adagrad~\cite{duchi2011adaptive}, the optimizer used in our model. In Adagrad, model parameters $\vw_{t}$ are updated by the rule
\begin{equation}\label{eq:adagrad}
\begin{aligned}
    \mG_t &= \mG_{t-1} + \vg_t^2, \\
    \vr_t &= \vg_t \cdot \mG_t^{-1/2}, \\
    \vw_{t+1} &= \vw_{t} - \eta_t \cdot \vr_t,
\end{aligned}
\end{equation}
where $\eta_t$ denotes the learning rate at step $t$, $\vg_t$ is the standard stochastic gradient of the empirical loss with respect to the model parameters and $\mG_t$, known as ``accumulator'', is a vector initialized to some small constant value, typically $0.1$. In addition, all powers operations are computed element-wise.

As mentioned in~\Secref{sec:understand}, we desire a more principled approach to control the step size when loss curvature is steep. However, the loss curvature measured by the eigenvalue of loss Hessian is very expensive to compute during training. Fortunately, the first-order gradients $\vg_t$ can be used as a surrogate for the Hessian (c.f.~\cite{Zhang2020Why}). 
Consequently, gradient clipping based algorithms become very popular to improve training stability and used in many large models~\cite{chowdhery2022palm, thoppilan2022lamda, brown2020language}.

\textbf{Gradient Clipping.} 
Proposed by~\citet{pascanu2013difficulty}, Gradient Clipping (GC) limits the magnitude of gradient (measured by its norm) before applying it to the model. In other words, as gradient magnitude becomes large (loss curvature becomes steeper), Gradient Clipping controls the ``effective step size'' to stabilize model training.

Formally, Gradient Clipping algorithm clips the gradients $\vg_t$ (before applying Adagrad update in~\eqref{eq:adagrad}) as:
\begin{equation}\label{eq:gc}
\begin{aligned}
\vg \rightarrow&
\left\{
	\begin{array}{ll}
		\lambda \frac{\vg}{\norm{\vg}}  & \mbox{if } \norm{\vg} \ge \lambda, \\
		\vg & \mbox{else}.
	\end{array}
\right. 
\\
\text{Or }\vg \rightarrow&~\sigma \cdot \vg, \textrm{ where } \sigma = \min \{ \frac{\lambda}{\norm{\vg}}, 1.0 \}
\end{aligned}
\end{equation}
The clipping threshold $\lambda$ is a hyperparameter that controls the maximum allowable gradient norm $\norm{\vg}$. In other words, if the model gradient $\vg_t$ has a large magnitude at step $t$, GC will clip its norm to $\lambda$ by rescaling gradients with a scalar \emph{clipping factor} $\sigma \in \sR^+$. In practice, Frobenius norm (or $L_2$ norm) $\norm{.}_2$ is a common choice for vector norm, and clipping is often applied to each layer independently of the other layers.

\textbf{Adaptive Gradient Clipping.}
Empirically, although GC can improve training stability of the model, training stability is extremely sensitive to the choice of the clipping threshold $\lambda$, requiring fine-grained tuning for different layers. What's worse, the threshold $\lambda$ need to be re-tuned when model structure, batch size, or learning rate is changed.

To overcome this burden, \citet{brock2021high} proposed Adaptive Gradient Clipping (AGC). AGC is motivated by the observation that the ratio of the norm of the gradients $\norm{\vg_t}$ to the norm of the model parameters $\norm{\vw_t}$ should not be large, otherwise training is expected to be unstable.

Specifically, the gradients $\vg$ is clipped by
\begin{equation}\label{eq:agc}
\begin{aligned}
\vg \rightarrow&
\left\{
	\begin{array}{ll}
		\lambda \frac{\norm{\vw}}{\norm{\vg}} \vg  & \mbox{if } \frac{\norm{\vg}}{\norm{\vw}} \ge \lambda, \\
		\vg & \mbox{else }.
	\end{array}
\right.
\\
\text{Or }\vg \rightarrow&~\sigma \cdot \vg, \textrm{ where } \sigma = \min \{ \lambda \frac{\norm{\vw}}{\norm{\vg}}, 1.0 \}
\end{aligned}
\end{equation}
Intuitively, if at step $t$ the gradient norm $\norm{\vg_t}$ is greater than a fraction of the parameter norm $\lambda \cdot \norm{\vw_t}$, AGC will clip the gradient norm to $\lambda \norm{\vw_t}$, by rescaling gradients with a scalar \emph{clipping factor} $\sigma \in \sR^+$.
AGC can be viewed as a special case of GC, where the clipping threshold $\lambda^{\textrm{GC}}$ is a function of model parameters $\lambda^{\textrm{GC}} = \lambda^{\textrm{AGC}} \norm{\vw}$.
So when using AGC, we don't need to fine tune $\lambda$ for different layers, this is where the ``adaptiveness'' comes from.

\subsection{Observations of Training Dynamics}
Despite the success of GC and AGC in various domains, we found that they are not effective enough to prevent loss divergence when being applied in our model. 
To better understand the limitations of GC/AGC and to propose better solutions, we inspect the training of our model without using any gradient clipping based techniques\footnote{Note the model we are inspecting here is the model-b in~\Figref{fig:divergence}}. 

\Figref{fig:train_loss_auc} shows the training loss and AUC for a particular binary classification task. To simplify the illustration, let's look mainly at the 3 most important training steps: step-a, step-b, and step-c\footnote{The specific training step numbers are: step-a=198.7k, step-b=198.8k, step-c=198.9k.}. 
As we can see, this model is training healthily before step-a: the loss is minimized and the AUC has increased rapidly. However, at step-b, the model's training loss started to diverge and AUC began to drop, though relatively unnoticeably. Finally, at step-c, this model was fully diverged with loss become large, and AUC dropped to 0.5.

In~\Figref{fig:norms_factors_top_hidden}(left), we take a closer look at some statistics for the \emph{top shared layer} to understand what happened as loss diverged. 
The gradient norm $\norm{\vg}_2$ is pretty consistent before step-a when model is healthy. 
Then it grew to a large value at step-b, suggesting the loss curvature is quite steep at that moment.
Since we didn't apply any model stability treatments, the model diverged completely at step-c and the gradient norm $\norm{\vg}_2$ became a small value.
This means that all pre-activations (values before applying nonlinear activation) at this layer already reach a state where gradients are extremely small\footnote{One typical example is the dying ReLU where most pre-activations are smaller than zero. It is worth noting that  other nonlinear activations also have regions where gradients are close to zero, so can suffer from the same issue.}, causing the loss divergence to become nonrecoverable.

\begin{algorithm}[t!]
\caption{Adagrad with Clippy}
\begin{algorithmic}[1]\label{alg:clippy}
\STATE \textbf{Input:} Parameter vector to optimize $\vw$; objective function $\mathcal{L}$; learning rate schedule $\eta_t$.
\STATE \textbf{Input:} Clippy hyperarameters: relative threshold $\lambda_\textrm{rel}$ and absolute threshold $\lambda_\textrm{abs}$.
\STATE Initialize parameter vector $\vw_{0}$.
\FOR{$t=0$ to $T-1$}
\STATE $\vg_t = \frac{\partial \mathcal{L}(\vw_{t})}{\partial \vw_{t}} \rightarrow$ obtain stochastic gradient.
\STATE $ \mG_{t} = \mG_{t-1} + \vg_t^2 \rightarrow$ update accumulator
\STATE $ \vr_t = \vg_t \cdot \mG_{t}^{-1/2} \rightarrow$ compute updates
\STATE $ \sigma_t = \min\{1.0, \min ( \frac{ \lambda_\textrm{rel} | \vw_{t} | + \lambda_\textrm{abs} }{ \eta_t * | \vr_t |} ) \} \rightarrow$ get clipping factor
\STATE $ \vw_{t+1} = \vw_{t} - \eta_t \sigma_t \vr_t \rightarrow$ apply rescaled updates
\ENDFOR
\STATE \textbf{Return:} $\vw_T$
\end{algorithmic}
\end{algorithm}

Knowing what happened, we fabricate how GC/AGC will react in this situation.
\Figref{fig:norms_factors_top_hidden}(left) plots the measurements of $\norm{\vg}_2$ (blue) and $\frac{\norm{\vg}_2}{\norm{\vw}_2}$ (orange) that are used to determine clipping factors in GC and AGC.
Not surprisingly, both measurements became larger at step-b. However, the relative scale of change for these measurements are different. 
$\frac{\norm{\vg}_2}{\norm{\vw}_2}$ (orange) is more sensitive to loss curvature changes than $\norm{\vg}_2$ (blue).
The difference in sensitivity of these measurements can result in different \emph{clipping factors} $\sigma$, which is the rescaling multiplier to the gradients in different methods.
\Figref{fig:norms_factors_top_hidden}(right) gives the clipping factor $\sigma$ for GC and AGC when using $\lambda^{\textrm{GC}}=10^{-1}$ and $\lambda^{\textrm{AGC}}=10^{-3}$ as clipping thresholds\footnote{As described in~\Secref{sec:experiments_baseline}, we obtain these thresholds with grid-search and a even lower threshold can impact convergence.}.

By checking the clipping factors, we hypothesize that the reason behind inefficacy of GC/AGC is that they failed to offer enough constraints on gradients (i.e., failed to provide enough control over the ``effective step size'') when gradient norm suddenly increases (i.e., loss curvature becomes steep),  due to lack of sensitivity.
More specifically, \textbf{both methods rely on $L_2$ norm, which is not sensitive to drastic gradient changes in only a few coordinates, especially when layer width is large}.

\subsection{Proposed Solution: Clippy}
To alleviate this limitation, we proposed a new algorithm called Clippy. Clippy has two major changes over GC/AGC: 
First, it uses $L_\infty$ norm instead of $L_2$ norm to increase its sensitivity to changes in individual coordinates.
Second, it clips based on updates $\vr_t=\vg_t \cdot \mG_t^{-1/2}$ instead of gradients $\vg_t$, since updates are the actual change to model parameters and can be quite different from gradients when using the Adagrad optimizer.

Specifically, Clippy controls 
\begin{equation}\label{eq:clippy}
\norm{\frac{\vr_t}{\vw_{t}}}_{\infty} < \lambda,
\end{equation}
and then rescales updates when the inequality is violated.
From~\Figref{fig:norms_factors_top_hidden}, we can see that this measurement has a a more dramatic change at step-b, when loss was diverging. 
Suppose we use $\lambda^{\textrm{Clippy}} = 10^{-1}$ as the clipping threshold, Clippy results in 2 orders of magnitude smaller clipping factors $\sigma$ compared to GC/AGC, thanks to the better sensitivity of the measurement.
In other words, we hope Clippy can put \textbf{larger constraints on the actual updates when loss curvature is steep even in a few coordinates}.

Formally, we present Clippy in~\Algref{alg:clippy}. As can be seen, there are some minor but important changes in line-8 of the algorithm compared to what we described in~\eqref{eq:clippy}.
\begin{enumerate}
\item \textbf{Introducing absolute threshold.}
In Clippy, we use two hyperparameters: The relative threshold $\lambda_\textrm{rel}$ that is similar to GC/AGC, and another absolute threshold $\lambda_\textrm{abs}$. With the absolute threshold $\lambda_\textrm{abs}$ introduced, we can avoid aggressive clipping when model parameters are zero (e.g., biases that are initialized to zeros) or have very small values. As will be discussed in~\Secref{sec:method_relation}, this allows Clippy to switch from GC-style to AGC-style during training.

\item \textbf{Considering learning rate.}
We have learning rate $\eta_t$ in the denominator when calculating the clipping factor to account for different learning rate schedules. If the learning rate slowly ramps up, this will loosen the clipping threshold at initial training, avoiding a slow pace of convergence in the initial phases of training.

\end{enumerate}

\subsection{Additional Discussions}
\subsubsection{Relationship with other methods}\label{sec:method_relation}
Clippy has interesting connections with other methods.
In gradient clipping based algorithms, if we accumulate the accumulator with original gradients (instead of clipped gradients). Then, we can have a general Adagrad update form with all aforementioned algorithms
\begin{equation}\label{eq:general_adagrad}
\begin{aligned}
    \vr_t &= \vg_t \cdot \mG_t^{-1/2}, \\
    \vw_{t+1} &= \vw_{t} - (\eta_t \sigma_t) \vr_t.
\end{aligned}
\end{equation}
That is, different algorithms scale down the learning rate $\eta_t$ with different choices of clipping factor $\sigma_t$.
And the choice of clipping factors by different algorithms are summarized in the table below.
\begin{table}[h!]
\begin{tabular}{r|c}
\toprule
\textbf{Algorithm} & $\sigma_t$\\
\toprule
GC~\cite{pascanu2013difficulty} & $\min \{1.0, \lambda \frac{1}{\norm{\vg}_2} \}$ 
\\
\midrule
AGC~\cite{brock2021high} & $\min \{ 1.0, \lambda \frac{\norm{\vw}_2}{\norm{\vg}_2} \}$
\\
\midrule
LAMB~\cite{you2019large} & $ \frac{\phi(\norm{\vw_{t}}_2)}{\norm{\vr_t}_2}$
\\
\midrule
Clippy~(Ours) & $\min\{1.0, \min ( \frac{ \lambda_\textrm{rel} | \vw_{t} | + \lambda_\textrm{abs} }{ \eta_t * | \vr_t |} ) \}$
\\
\bottomrule
\end{tabular}
\label{tb:relation}
\end{table}

\noindent{Clippy is a combination of  GC/AGC/LAMB:} 
First of all, Clippy switches from GC-style to AGC-style during training. During initial model training when $|\vw| \approx 0$, $\lambda_\textrm{abs}$ dominates the clipping threshold $\frac{ \lambda_\textrm{rel} | \vw_{t} | + \lambda_\textrm{abs} }{ \eta_t * | \vr_t |} \approx \frac{ \lambda_\textrm{abs} }{ \eta_t * | \vr_t |}$ and makes Clippy close to GC. In later training, when $\lambda_\textrm{rel} |\vw| \gg \lambda_\textrm{abs}$, Clippy acts more like AGC $\frac{ \lambda_\textrm{rel} | \vw_{t} | + \lambda_\textrm{abs} }{ \eta_t * | \vr_t |} \approx \frac{ \lambda_\textrm{rel} | \vw_{t} |}{ \eta_t * | \vr_t |}$.
However, compared to GC/AGC, Clippy relies on updates instead of gradients.
Moreover, although both Clippy and LAMB use the updates, Clippy does not completely ignore the update magnitude as in LAMB\footnote{LAMB updates parameter by $\vw_{t+1} = \vw_{t} - \eta_t \frac{\vr_t}{\norm{\vr_t}_2}\phi(\norm{\vw_{t}}_2)$, with $\phi(x)=\min\{\max\{x, \gamma_l \}, \gamma_u \}$ bounds the parameter $L_2$ norm. It uses only the direction of updates and ignores its magnitude.}.
Finally, Clippy uses $L_\infty$ instead of $L_2$ norm to be more sensitive to drastic update changes in a small number of coordinates.

\subsubsection{Clip locally or globally}\label{sec:method_relation_local}
When using Clippy, we clip the update per each layer (\emph{a.k.a} locally) instead of per all model parameters as a whole (\emph{a.k.a} globally), similar to the other methods (like GC/AGC/LAMB).
This gives more flexibility on finer-grained control, but results in a biased gradient update. However, in large-batch settings, it can be shown that this bias is small~\cite{you2019large}.

\subsubsection{Adapting to other optimizers}
One can easily adapt Clippy to optimizers other than Adagrad by using the optimizer-dependent update $\vr_t$. Empirically, we have also observed clear benefits in training stability when applying Clippy on Adam~\cite{kingma2014adam}, without compromising convergence. But we leave the theoretical convergence analysis of Clippy to future work.

\section{Empirical Studies}\label{sec:experiments}
Conducted on a YouTube production dataset, experiments in this section are divided into two parts. Firstly, we compare Clippy with other baselines to verify its benefits for improving model stability. Then we show some further analyses for Clippy to better understand its strength.

\subsection{Experiment Setup}

\begin{table}[t!]
\caption{The configuration of each model setting.}
\begin{tabular}{c|c|c}
\toprule
\multirow{2}{*}{\textbf{Model Name}} & \textbf{Non-Embedding} & \textbf{Shared bottom}\\
 & \textbf{Model Parameters} & \textbf{Architecture}\\
\toprule
\texttt{Small} &  7.5M &  FFN: $512 \times 2$\\
\midrule
\texttt{Large} & 57.0M & FFN: $4096 \times 4$\\
\midrule
\texttt{Large+DCN} & 68.0M  & $\textrm{DCN + LN} \rightarrow 4096 \times 4$\\
\bottomrule
\end{tabular}
\label{tb:model_config}
\end{table}

\subsubsection{Model detail.} 
Besides all the model properties that are already covered in~\Secref{sec:understand_model}, it is worth mentioning that we simplified our ranking model by (1) Only keeping the most important subset of tasks and input features; (2) Using a simple shared bottom structure with several shared hidden layers.
Though much simpler than the production model, we found it to be a sufficiently good testbed for studying the training stability problem, as it allows us to train models faster and focus more on research perspectives instead of irrelevant modeling details.
The model is built with TensorFlow-2~\cite{abadi2016tensorflow} and is trained using a large batch size of ~65k on TPUs.

\begin{table*}[t!]
\begin{tabular}{c|c|ccccc}
\toprule
\multirow{2}{*}{\textbf{Model Name}} & \multirow{2}{*}{\textbf{Metrics}} & \multicolumn{5}{c}{\textbf{Methods}}\\
& & \textbf{Naive} & \textbf{GC} & \textbf{AGC} & \textbf{LAMB} & \textbf{Clippy} \\
\toprule
\multirow{3}{*}{\texttt{Small}} & AUC {\scriptsize(higher is better}) 
& \multirow{3}{*}{\textit{diverged}} & 71.68~{\scriptsize $\pm$~0.13} & 71.73~{\scriptsize $\pm$~0.00} & 71.56~{\scriptsize $\pm$~0.01} & \underline{71.79}~{\scriptsize $\pm$~0.00} \\
& RMSE {\scriptsize(lower is better)} & & 1.058~{\scriptsize $\pm$~0.002} & 1.059~{\scriptsize $\pm$~0.003} & 1.063~{\scriptsize $\pm$~0.001} & \underline{1.056}~{\scriptsize $\pm$~0.000}\\
& Best learning rate & & 2x & 1x & 1x & 2x\\
\midrule

\multirow{3}{*}{\texttt{Large}} & AUC {\scriptsize(higher is better}) 
& \multirow{3}{*}{\textit{diverged}} & 72.07~{\scriptsize $\pm$~0.05} & 72.09~{\scriptsize $\pm$~0.02} & 72.01~{\scriptsize $\pm$~0.09} & \underline{72.16}~{\scriptsize $\pm$~0.02} \\
& RMSE {\scriptsize(lower is better)} & & 1.053~{\scriptsize $\pm$~0.003} & \underline{1.051}~{\scriptsize $\pm$~0.001} & 1.054~{\scriptsize $\pm$~0.002} & \underline{1.051}~{\scriptsize $\pm$~0.000}\\
& Best learning rate & & 2x & 1x & 1x & 2x\\
\midrule

\multirow{3}{*}{\texttt{Large+DCN}} & AUC {\scriptsize(higher is better}) 
& \multirow{3}{*}{\textit{diverged}} & 72.27~{\scriptsize $\pm$~0.03} & 72.06~{\scriptsize $\pm$~0.08} & 72.05~{\scriptsize $\pm$~0.11} & \underline{72.37}~{\scriptsize $\pm$~0.01} \\
& RMSE {\scriptsize(lower is better)} & & 1.049~{\scriptsize $\pm$~0.001} & 1.051~{\scriptsize $\pm$~0.001} & 1.057~{\scriptsize $\pm$~0.001} & \underline{1.047}~{\scriptsize $\pm$~0.001}\\
& Best learning rate & & 1x & 2x & 1x & 2x\\

\bottomrule
\end{tabular}
\caption{Evaluation of training stability treatments on different model settings. Methods suffering from training instability problems should get worse evaluation metrics. We first find the best learning rate (1x or 2x) for each variant, then repeat the same setting 3 times and report mean and standard deviation. We use \underline{underline} to denote the best result for each setting.}
\label{tb:overall_performance}
\end{table*}

\subsubsection{Evaluation protocol.} 
Unfortunately, there is no reliable metric to quantify the model's training stability. 
To precisely measure the benefit from better training stability, we vary model complexity as well as learning rates, then check model’s offline quality, measured by AUC for binary classification tasks and RMSE for regression tasks.
Presumably, a more complex model gives better offline quality but is more likely to suffer from loss divergence issues. So if an algorithm can significantly improve the model's training stability, we should observe better offline metrics when using it.
More specifically, we used first ($N-1$) days of data to sequentially train the model and continuously evaluate the model's performance (AUC or RMSE) on the last day (the $N$-th day) of data. 
If the model does not suffer from any loss divergence issues during training, we should observe the evaluation metrics keep becoming better, as the model is adapting to the data distribution closer to the $N$-th day of data.
Whereas if the model's loss diverges during training, either fully-diverge or consistently micro-diverge, the evaluation metrics will be significantly impacted.

To explore the effect of model complexities, we consider various model settings summarized in Table~\ref{tb:model_config}. 
Both \texttt{Small} and \texttt{Large} use simple feed-forward networks as the shared bottom, with two 512 layers and four 4096 layers respectively. \texttt{Large+DCN} is built on top of \texttt{Large} and further adds more complexity by having DCN-v2 layers~\cite{wang2021dcn} on inputs, followed by a standard Layer Normalization~\cite{ba2016layer}.

\begin{figure*}[t!]  %
\centering
    \centering
    \begin{subfigure}[b]{0.9\textwidth}
        \includegraphics[width=\textwidth]{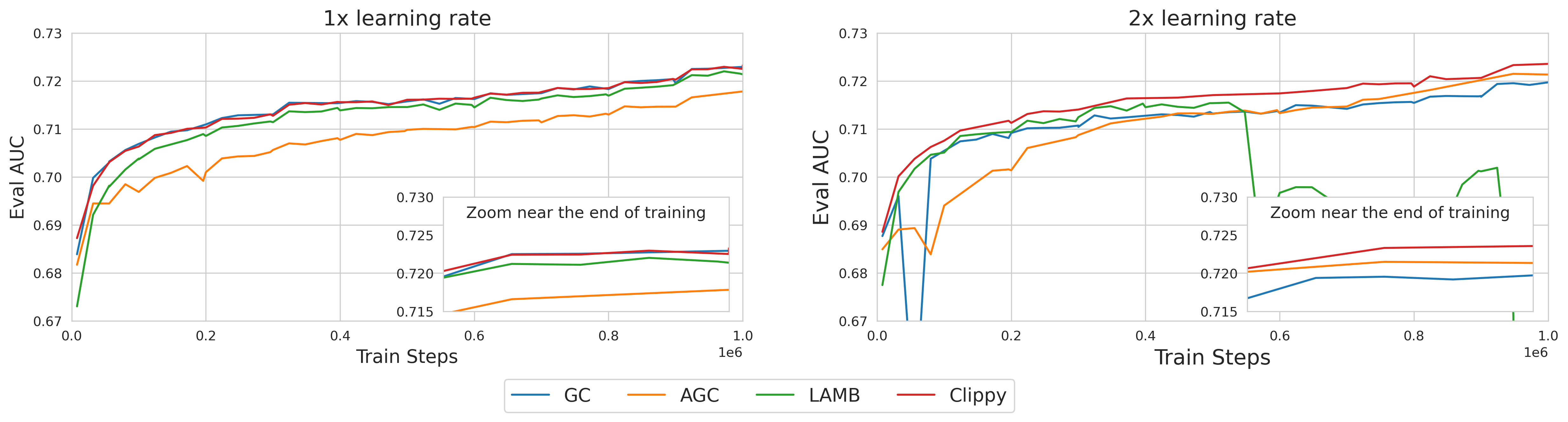}
        \vspace{-2em}
        \caption{AUC for different methods during the training on \texttt{Large+DCN} model.}
        \label{fig:eval_large_dcn}
    \end{subfigure}
    
    \begin{subfigure}[b]{0.9\textwidth}
        \includegraphics[width=\textwidth]{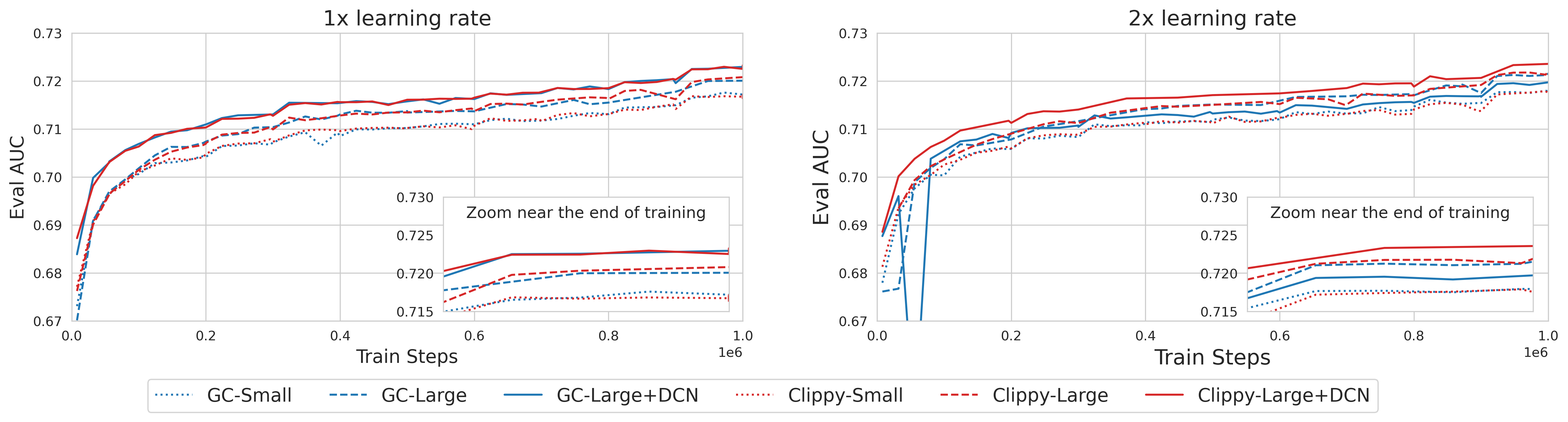}
        \vspace{-2em}
        \caption{AUC of Clippy and GC (the best baseline) during the training on different model settings.}
        \vspace{-1em}
        \label{fig:eval_gcclippy}
    \end{subfigure}

\caption{Evaluation AUC vs. Training steps for different methods in different model settings.}
\label{fig:eval_auc}
\end{figure*}

\begin{figure*}[t!]
\centering
\includegraphics[scale=0.4]{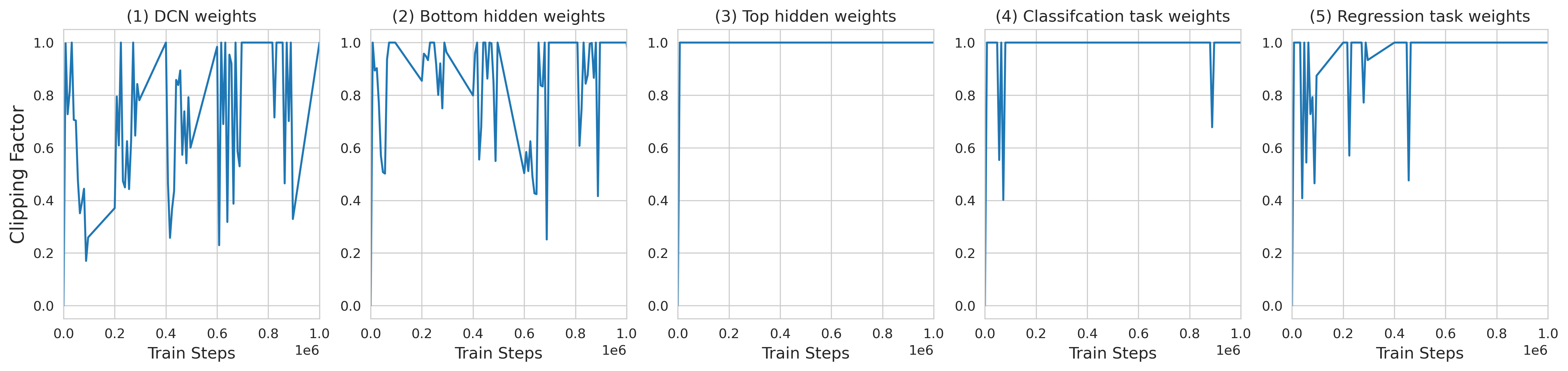}
\vspace{-1em}
\caption{Clippy's clipping factor on different layers during training the \texttt{Large+DCN} model.}
\label{fig:clippy_clipping_factor}
\end{figure*}

\subsubsection{Baselines.}\label{sec:experiments_baseline}
We apply Clippy and other baselines to non-embedding model parameters and compare their effectiveness. Below are more details about these baselines and Clippy.

\begin{itemize}
\item \textbf{Gradient Clipping (GC)}~\cite{pascanu2013difficulty}:
We used layer-wise (local) gradient clipping with clipping threshold searched from $\lambda^{\textrm{GC}} \in \{ \underline{10^{-1}}, 10^{-2}, 10^{-3}\}$.

\item \textbf{Adaptive Gradient Clipping (AGC)}~\cite{brock2021high}:
We used the official implementation\footnote{\url{https://github.com/deepmind/deepmind-research/tree/master/nfnets}} provided in the paper and searched clipping threshold from $\lambda^{\textrm{AGC}} \in \{ 10^{-2}, \underline{10^{-3}}, 10^{-4}\}$.

\item \textbf{LAMB (adapt to Adagrad)}~\cite{you2019large}:
LAMB was originally proposed based on Adam~\cite{kingma2014adam}, while the authors also provided a general form for the clipping which we introduced in~\Secref{sec:method_relation}. We choose $\phi(x)=x$ as in the official implementation\footnote{\url{https://github.com/tensorflow/addons/blob/master/tensorflow_addons/optimizers/lamb.py}}. Since LAMB uses parameter $L_2$ norm $\norm{\vw}_2$ as update magnitude that is different from other methods, we have to scale the learning rates by $\mu$ and searched $\mu \in \{ 10^{-1}, 10^{-2}, \underline{10^{-3}} \}$.

\item \textbf{Clippy}:
Clippy has two hyperparameters $\lambda_{\textrm{abs}}$ and $\lambda_{\textrm{rel}}$ so suppose to be more non-trivial for the tunings, but we found simply setting $\lambda_{\textrm{rel}}=0.5$ and $\lambda_{\textrm{abs}}=10^{-2}$ gives decent performance in our experiments. .
\end{itemize}

\subsection{Overall Performance}

Table-\ref{tb:overall_performance} presents the overall comparison between Clippy and other baselines on different model settings. 
Though the model is trained on six tasks, due to space limitations, we only present the metrics from two most representative tasks --- one binary classification task evaluated with AUC (in percentage) and another regression task evaluated with RMSE.
We not only use the original learning rate but also try to double the learning rate and see if any method can benefit from it.
After finalizing the best learning rate, we repeat the same setting 3 times with different random seeds and report the mean and standard deviation.

Looking at Table~\ref{tb:overall_performance}, we can see the naive method which does not have any treatments on training stability always suffers from loss divergence, even on the \texttt{Small} model. There is a chance for it to survive if we drastically tune down the learning rate (see~\Secref{sec:suppl_more_evidence} in Supplementary Material) but we omit its results here as they are bad.
GC can survive with 2x learning rate and provide good results on \texttt{Small} and \texttt{Large} model. But in a more complex model with DCN, GC can only use 1x learning rate, otherwise it will suffer from loss divergence issues (see blue line in ~\Figref{fig:eval_large_dcn} right).
AGC did a reasonable job on \texttt{Small} and \texttt{Large} with 1x learning rate, but became bad with 2x learning rate. On \texttt{Large+DCN}, AGC shows very high variance using either 1x or 2x learning rate (see orange line in~\Figref{fig:eval_large_dcn}), suggesting AGC already reaches its limits on keeping training stable.
LAMB successfully trains the model without suffering from training instability problems using 1x learning rate, but the convergence is negatively impacted. On~\Figref{fig:eval_large_dcn}, we found the results from LAMB are always worse than the other methods. We believe this is due to LAMB completely ignoring the update magnitude, causing the convergence at initial training to be very slow when parameter $L_2$ norm is small.
Surprisingly, GC performs the best on all settings among all the baselines, this could be because the model is relatively simple thus tuning the clipping threshold for GC is still trivial.

On the last column of Table~\ref{tb:overall_performance}, we can see Clippy handles all model settings with 2x learning rate. More importantly, Clippy doesn't compromise convergence, it has comparable results with GC (\emph{i.e.,} the best baseline) on \texttt{Small} and \texttt{Large} model (see~\Figref{fig:eval_gcclippy}), and having significantly better AUC (Note 0.1\% AUC improvement in our model is considered very significant and can lead to live metric gains) and RMSE on \texttt{Large+DCN} model compared to GC. One important finding we want to highlight is that Clippy offers larger gains when the model is more complex and trained with a larger learning rate. On~\Figref{fig:eval_gcclippy}, we can see gap between Clippy and GC is getting larger when using a more complex model with 2x learning rate.
So we are not surprised Clippy can help in the production model which is much more complex than \texttt{Large+DCN}.

\subsection{Closer Look at Clippy's Clipping Factors}
\Figref{fig:clippy_clipping_factor} shows Clippy's clipping factor on different layers during training the \texttt{Large+DCN} model.
As introduced in~\Secref{sec:method}, the clipping factor $\sigma \in (0.0, 1.0]$. A smaller clipping factor indicates more clipping is done to scale down the learning rate.
Since the clipping is applied per layer, we plot the clipping factor for several layers, including the weights in (1) DCN layer, in (2) top and (3) bottom hidden layer of shared bottom, and in the output layers of (4) binary classification task and  (5) regression task.
It is interesting to see more clipping is done on the bottom layers of the model. We think this intuitively makes sense, because bottom layers usually have smaller parameter norm so Clippy's clipping threshold will also be smaller. On the other hand, this could potentially benefit training stability because we know a small change in the bottom layer weights can lead to a large difference in model outputs.

\section{Conclusion and Future Work}\label{sec:conclusion}
In this paper, we present the training instability problem that happen recurrently in ranking models at YouTube. We show the importance and challenges of mitigating this problem in the long run. To better understand the issue, we dive deep into the problem, trying to know the root cause of why conventional methods did not work well in our case. With our understandings, we propose a new clipping based method called Clippy, which has a nice relationship with existing methods but alleviates the limitations of them. From empirical studies on the YouTube production dataset, we found Clippy showed significantly better strength on improving model training stability than other baselines.

Clippy showed significant improvements on training stability in multiple ranking models for YouTube recommendations. It is productionized in some large and complex models. More importantly, it unblocks several ongoing modeling developments and alleviates us from training instability problems that happened recurrently.

As for future work, we hope to theoretically justify the effectiveness of Clippy and provide convergence guarantees. In addition, we hope evolution-based AutoML algorithms~\cite{chen2022evolved,so2021primer} can be applied here to do a better job than the human-designed clipping methods.

\newpage
\bibliographystyle{ACM-Reference-Format}
\balance
\bibliography{references}

\newpage
\appendix

\begin{table*}[t!]
\begin{tabular}{rc|c|c|c}
\toprule
\textbf{Direction} & \textbf{Specific Change} & \textbf{\#Diverged} & \textbf{\#Tried} & \textbf{Divergence Ratio}\\
\toprule
N/a & Clean & 5 &  5 & 100\%\\
\midrule
Model size  & Smaller model size &  1 &  3 & 33\%\\ 
Model complexity & Remove DCN &  2 &  3 & 66\%\\
\midrule
Input features & Remove subset of input features &  2 &  3 & 66\%\\
\multirow{2}{*}{Output tasks} & Remove one task &  2 &  3 & 66\%\\
 & Remove two tasks &  0 &  3 & 0\%\\
\bottomrule
\end{tabular}
\caption{The impact of different model properties to training instability. \texttt{Large+DCN} model is used without any treatments on training stability and with 0.4x learning rate.}
\label{tb:more_studies}
\end{table*}

\begin{figure}[t!]  %
    \centering
    \includegraphics[width=0.45\textwidth]{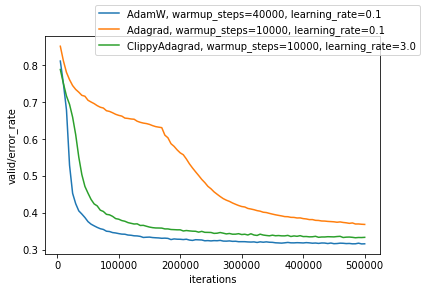}
\caption{A comparison of AdamW, Adagrad and Adagrad with Clippy on the task for English to German translation.}
\label{fig:xformer_comparison}
\end{figure}

\begin{figure*}[t!]  %
\centering
    \centering
    \begin{subfigure}[b]{0.95\textwidth}
        \includegraphics[width=\textwidth]{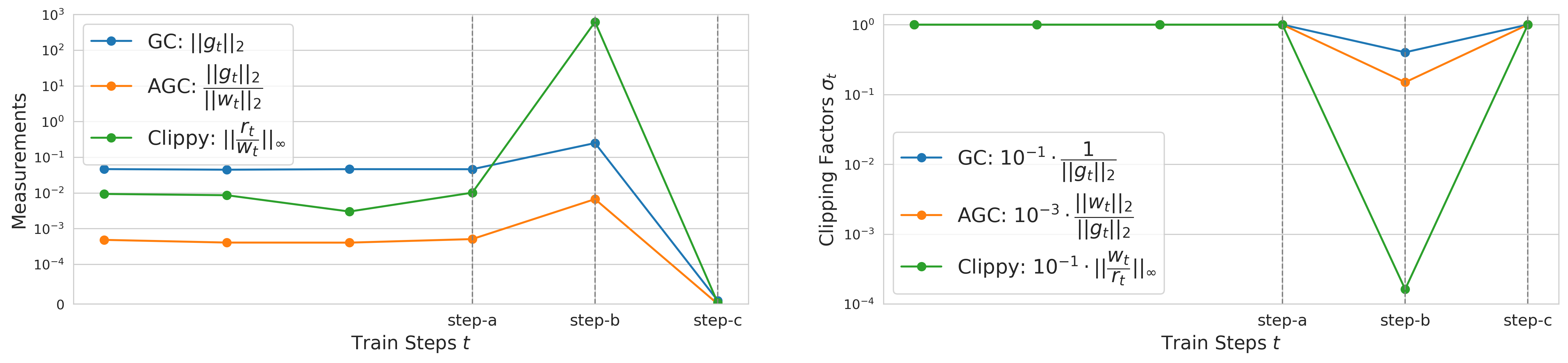}
        \caption{Bottom hidden layer weights.}
        \label{fig:norms_factors_bottom_hidden}
    \end{subfigure}
    
    \begin{subfigure}[b]{0.95\textwidth}
        \includegraphics[width=\textwidth]{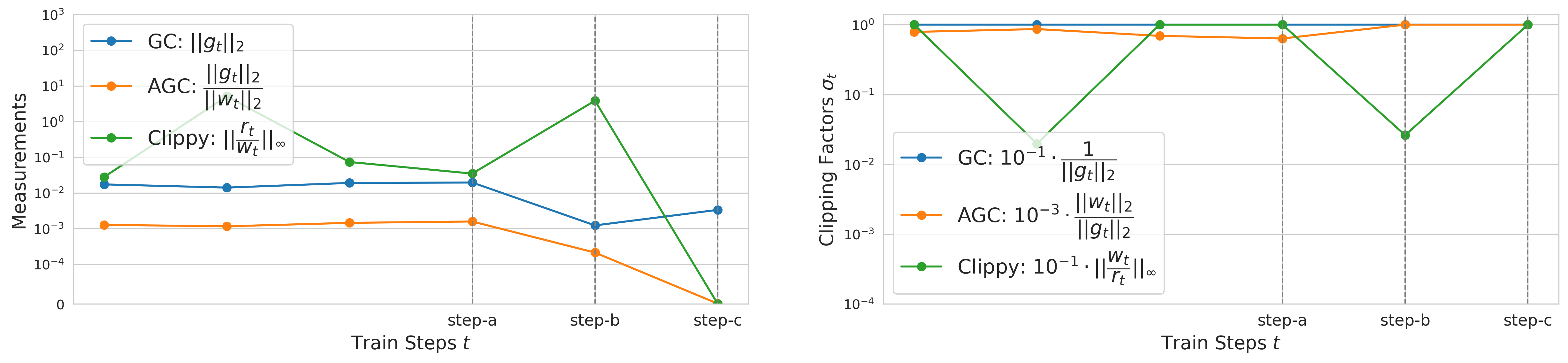}
        \caption{Classification task layer weights.}
        \label{fig:norms_factors_cls}
    \end{subfigure}
    
    \begin{subfigure}[b]{0.95\textwidth}
        \includegraphics[width=\textwidth]{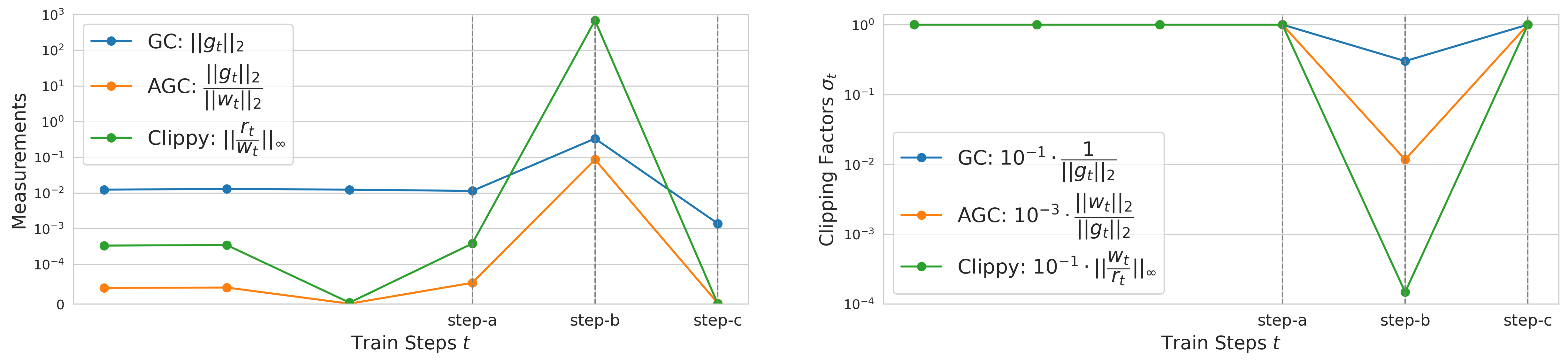}
        \caption{Regression task layer weights.}
        \label{fig:norms_factors_reg}
    \end{subfigure}

\caption{(left) Measurements used by different methods to determine clipping factors. (right) The corresponding clipping factors.}
\label{fig:training_statistics_extension}
\end{figure*}

\section{Additional empirical studies}

\subsection{Empirical Evidence for~\Secref{sec:understand_cause}}\label{sec:suppl_more_evidence}
In Table~\ref{tb:more_studies}, we show more empirical evidence on how the unique properties of multitask ranking models in the recommendation domain can affect training stability. All the experiments are performed on the \texttt{Large+DCN} model without any treatments on training stability. We used 0.4x learning rate to make the training at the edge of instability and to see the benefits from other changes on improving training stability.

As a result, we can see that the loss of the model kept diverging if no change is applied: 5 out of 5 runs has loss fully diverged. However, if we reduce the model size (by switching to \texttt{Small+DCN} model) or remove DCN-v2 layers (by switching to \texttt{Large} model), there will be some surviving cases. Moreover, if we remove a subset of input features or one or more output tasks, we can also observe their benefits on training stability. Due to the large training cost for each trial, we cannot offer more data points, but we hope these results can support for the hypotheses and claims in~\Secref{sec:understand_cause}.

\subsection{A Transformer-based model}
In this section, we report the performance of Clippy in an additional setting.
We based our experiment on init2winit\footnote{\url{https://github.com/google/init2winit}}'s~\cite{init2winit2021github} `translate\_wmt' dataset with the default `xformer\_translate' model, containing six encoder and six decoder layers for the task of English to German translation. We compared the default AdamW optimizer to Adagrad and to Adagrad with Clippy. For both AdamW and Adagrad optimizers we tested the learning rates [0.01, 0.03, 0.1, 0.3, 1.0], while for Adagrad with Clippy we used learning rate in [0.1, 0.3, 1.0, 3.0, 10.0, 30.0] and set $\lambda^{\textrm{GC}} = 0.1$, $\lambda^{\textrm{AGC}}=10^{-3}$.
All experiments were executed twice for 500k steps, once a with warm-up period 10k steps and a second time with a 40k step warm-up period.

When the warm-up period was 10k steps, AdamW diverged when the learning rate was 0.1 and diverged for learning rate equal to 1.0 when the warm-up period was at 40k steps. Adagrad diverged for learning rate equal to 0.3 in both cases.
On the other hand, Adagrad with Clippy did not diverge for any of our experiments. In a few of our earlier trials, Adagrad with Clippy did start to slowly diverge, however, that divergence was transitory and the model later recovered.

The baseline AdamW optimizer attained the best result with a learning rate of 0.1 and 40k warm-up period, reaching a validation error rate of 31.6\%. Adagrad's best run was with a learning rate of 0.1 and 10k warm-up period, reaching a validation error rate of 36.9\%, while Adagrad with Clippy's best run was with learning rate of 3.0 and 10k warm-up steps, reaching an error rate of 33.4\% on the validation set. See~\Figref{fig:xformer_comparison} for the validation error rate throughout the training process.

Note that although Adagrad with Clippy did not reach the performance of AdamW, which is considered state-of-the-art for this task, it did show significant improvement over the Adagrad implementation. Furthermore, we did not attempt to tune any of the model's parameters beyond what is reported above, opening the way to further improvements.

\section{Statistics from other layers}\label{sec:suppl_statistics}

Besides the top hidden layer weights presented in~\Figref{fig:norms_factors_top_hidden}, we also show statistics from other representative layers in~\Figref{fig:training_statistics_extension}. From the figure, we can see other layers behave similarly as the top hidden layer at step-b, except for the binary classification layer. From~\Figref{fig:norms_factors_cls}, we found measurements used by GC/AGC even failed to capture the sudden changes in model parameters at step-b, resulting in no clipping applied at step-b for the binary classification layer weights by GC/AGC.

\end{document}